\documentclass[11pt]{article}

\usepackage[final]{acl}

\usepackage{times}
\usepackage{latexsym}

\usepackage[T1]{fontenc}

\usepackage[utf8]{inputenc}

\usepackage{microtype}

\usepackage{inconsolata}

\usepackage{graphicx}

\usepackage{amsmath,amssymb,amsfonts,mathtools,bm}
\usepackage[ruled,vlined,linesnumbered]{algorithm2e}
\usepackage{xcolor}
\usepackage{booktabs}
\usepackage{multirow}
\title{VLA-Forget: Vision-Language-Action Unlearning for Embodied Foundation Models}

\author{Ravi Ranjan\thanks{Corresponding author.\\Accepted to the \textbf{ACL-2026} KnowFM, \\To appear in  ACL Proceedings.\\Code: \url{https://github.com/raviranjan-ai/VLA-Forget}}\\
  Florida International University\\
  Miami, USA \\
  \texttt{rkuma031@fiu.edu} \\\And
  Agoritsa Polyzou \\
  Florida International University\\
  Miami, USA \\
  \texttt{apolyzou@fiu.edu} \\}

\begin{document}
\maketitle
\vspace{-0.6cm}
\begin{abstract}
\vspace{-0.2cm}
Vision-language-action (VLA) models are emerging as embodied foundation models for robotic manipulation, but their deployment introduces a new unlearning challenge: removing unsafe, spurious, or privacy-sensitive behaviors without degrading perception, language grounding, and action control. In OpenVLA-style policies, behavior is produced through a fused visual encoder, a cross-modal projector, and a language backbone that predicts tokenized robot actions, so undesirable knowledge can be distributed across perception, alignment, and reasoning/action layers rather than confined to a single module. Consequently, partial unlearning applied only to the vision stack or only to the language backbone is often insufficient, while conventional unlearning baselines designed for standalone vision or language models may leave residual forgetting or incur unnecessary utility loss in embodied settings. We propose \textbf{VLA-Forget}, a hybrid unlearning framework that combines ratio-aware selective editing for perception and cross-modal specificity with layer-selective reasoning/action unlearning for utility-preserving forgetting. VLA-Forget jointly optimizes three objectives: targeted forgetting, perceptual preservation, and reasoning retention, through staged updates over the visual encoder, projector, and upper action-generating transformer blocks. 
Across forget-set behavior probes and retain-task evaluations, VLA-Forget improves forgetting efficacy by \textbf{10\%}, preserves perceptual specificity by \textbf{22\%}, retains reasoning and task success by \textbf{9\%}, and reduces post-quantization recovery by \textbf{55\%} relative to strong unlearning baselines.
\end{abstract}

\begin{figure}[t]
   \vspace{-0.4cm}
   \includegraphics[width=\linewidth]{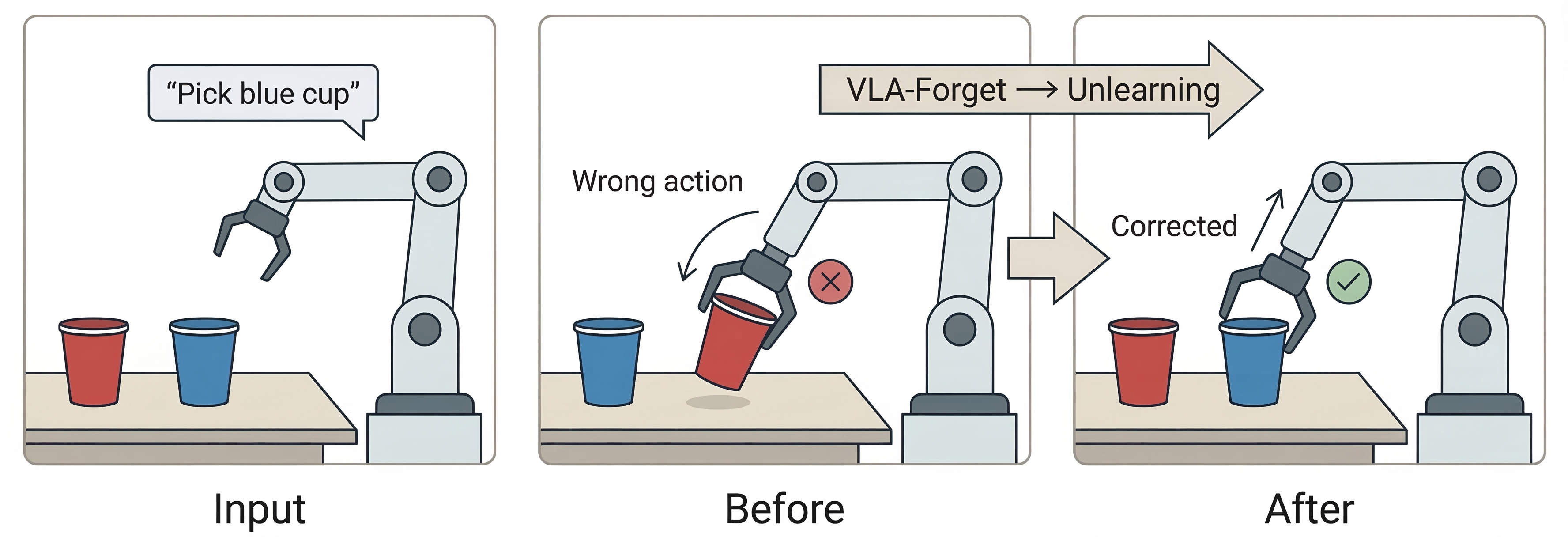}
    \caption{\textbf{Motivating failure case for VLA-Forget.} Given the instruction ``Pick blue cup,'' the policy produces a visually plausible but semantically incorrect action by grasping the red cup. After targeted unlearning, the policy suppresses this spurious instruction-to-action association and recovers instruction-consistent behavior.}
   \label{fig:intro}
   \vspace{-0.4cm}
\end{figure}

\vspace{-0.4cm}
\section{Introduction}
\label{sec:intro}
\vspace{-0.2cm}
\noindent Vision-Language-Action (VLA) models are emerging as a practical route to \emph{embodied foundation models}: policies that translate natural-language instructions and raw visual observations directly into robot actions by leveraging large pretrained vision and language backbones. OpenVLA exemplifies this paradigm as an open 7B VLA trained on 970K real-world robot demonstrations from Open X-Embodiment, combining a fused (DINOv2+SigLIP) visual encoder with a Llama~2 backbone and an action-token prediction interface that enables scalable fine-tuning and deployment on commodity hardware \cite{kim2024openvla,o2024open,pang2025openvla}.

\noindent However, \emph{deployment} of VLA policies introduces an urgent governance and reliability problem that is not fully addressed by today’s training and evaluation pipelines: once a robot policy has been trained, it may retain (i) unsafe or undesirable behaviors from a subset of demonstrations, (ii) privacy or IP-sensitive content embedded in multimodal data, or (iii) spurious shortcuts that appear correct under standard benchmarks but fail under structured distribution shift. The consequences of such failures are amplified in robotics because errors translate into physical actions. Recent evidence further highlights a VLA-specific grounding failure, sometimes described as \emph{linguistic blindness} where policies execute visually plausible trajectories even when the instruction is semantically contradictory, motivating stronger mechanisms for selective behavior removal and reliable grounding diagnostics \cite{zhuang2026q}. 

\noindent Figure~\ref{fig:intro} illustrates the core VLA-specific challenge studied in this work: undesirable behavior may arise from misalignment across perception, cross-modal grounding, or action priors, causing the policy to execute a physically plausible yet instruction-inconsistent action. This motivates \emph{selective unlearning} that removes a targeted erroneous behavior slice while preserving normal scene understanding and non-target task execution. Motivation is further supported by recent evidence that VLA policies remain brittle under visual corruption and grounding perturbations, underscoring the need for post hoc correction mechanisms beyond standard fine-tuning~\cite{lin2025failsafe,orjuela2026improving}.

\noindent Machine unlearning provides a practical way to remove targeted training influence without full retraining, but VLA policies are harder to unlearn than standard vision or language models. Unlike static predictors, a VLA model is a \emph{closed-loop control policy}, so failure must be judged through embodied behavior rather than output accuracy alone. In OpenVLA, continuous robot actions are represented as \emph{discrete action tokens}, which means unwanted behavior can be encoded jointly in visual features, cross-modal alignment, and instruction-conditioned action priors in the language backbone \cite{kim2024openvla}.

\noindent As a result, unlearning only the vision stack or only the language model is often insufficient: removing a visual trigger may leave the downstream action prior intact, while editing language priors may preserve harmful perceptual shortcuts or incorrect visual-language bindings. Existing baselines were largely designed for unimodal settings and transfer only partially to VLA policies. For example, exact approaches such as SISA require training-time changes \cite{bourtoule2021machine}, while approximate methods such as SCRUB improve forget-retain trade-offs but do not directly address multimodal component entanglement or control-oriented evaluation \cite{kurmanji2023towards}. Moreover, VLA unlearning involves competing retain, forget, and mismatch objectives over large backbones, motivating gradient-conflict mitigation such as PCGrad \cite{yu2020gradient}.

\noindent In this work, we introduce \textbf{VLA-Forget}, a hybrid unlearning framework for Vision-Language-Action models that is explicitly \emph{component-aware} and \emph{deployment-oriented}. VLA-Forget (i) targets perception and cross-modal specificity through ratio-aware selective editing of the visual encoder and projector, (ii) targets reasoning/action utility-preserving forgetting through significance-based selective editing of action-relevant transformer blocks, and (iii) performs staged, adapter-first updates (e.g., LoRA) to enable efficient unlearning with rollback and canary deployment compatibility \cite{hu2022lora,kim2025fine,liu2025hybridvla}. We evaluate unlearning with robotics-centric metrics and unlearning-centric audits (including safety violation risk); furthermore, we evaluate the model's robustness post-quantization, aligning the evaluation with the realities of VLA deployment~\cite{carlini2022membership,zang2025rlinf, zhang2024catastrophic}
\noindent While VLA-Forget improves targeted behavior suppression in benchmarked embodied settings, it is an approximate unlearning method and should not be interpreted as providing certified erasure.

\noindent \textbf{Key Contributions.} \textbf{(i)} We formalize \emph{VLA unlearning} as a three-goal problem \textbf{targeted forgetting} (efficacy), \textbf{perceptual preservation} (specificity), and \textbf{reasoning retention} (utility) in the presence of an action-token interface and multimodal component entanglement. 
\textbf{(ii)} We propose \textbf{VLA-Forget}, a novel hybrid unlearning pipeline that exploits ratio-aware selective editing for perception/cross-modal specificity with significance-based selective editing for reasoning/action utility preservation, implemented in an adapter-first manner compatible with OpenVLA fine-tuning workflows. 
\textbf{(iii)} We outline an \textbf{evaluation protocol} that couples embodied performance (task success and control stability) with unlearning audits (forget/retain scores and safety violation risk), and incorporates structured contradiction probes to diagnose and prevent spurious “fake success” under OOD instructions. 

\section{Related Work}
\label{sec:related-work}

Vision-language-action (VLA) models extend multimodal foundation models from perception and generation to embodied control. Early systems such as VIMA and RT-2 showed that robot behavior can be conditioned on interleaved visual-language prompts and represented through tokenized actions, while Open X-Embodiment and OpenVLA scaled this paradigm to diverse real-robot data and open 7B policies. Despite this progress, prior VLA research has focused mainly on scaling, transfer, and adaptation, rather than post hoc removal of unsafe behaviors, undesirable concepts, or sensitive instruction-action associations. \cite{zhu2023vima,zitkovich2023rt,o2024open,grover2026embodied,kim2024openvla}

\noindent Machine unlearning has progressed from exact retraining-based deletion to approximate updates that balance erasure quality, efficiency, and retained utility. In language models, recent work spans gradient-ascent, retain-regularized, preference-based, and activation-space methods, while emphasizing joint evaluation of forgetting, utility preservation, and privacy leakage. Two recurring issues are especially relevant for embodied policies: broad model-wide edits often cause collateral degradation, and apparently successful forgetting can fail under deployment transformations such as quantization. These limitations motivate selective, structure-aware unlearning over indiscriminate full-model updates. \cite{bourtoule2021machine,maini2024tofu,yao2024machine,liu2025rethinking,zhang2024catastrophic}

\noindent Related work in vision, vision-language, and diffusion models studies forgetting of classes, identities, and concepts in multimodal transformers. SSD and SalUn localize updates through parameter importance or saliency, LoTUS improves scalable approximate unlearning with uncertainty-aware smoothing, and SLUG shows that targeted single-layer editing can sometimes suffice. In generative settings, ESD and UCE edit text-conditioned diffusion behavior at the concept level. Collectively, these methods expose a central trade-off: aggressive updates improve forgetting but may damage retention, whereas overly localized edits can fail when unwanted knowledge is distributed across layers and modalities. \cite{foster2024fast,fan2023salun,spartalis2025lotus,cai2024targeted,gandikota2023erasing,gandikota2024unified}

\noindent VLA unlearning differs from prior LLM- or VLM-only settings because undesired behavior may be encoded jointly in visual features, cross-modal bindings, and action-token priors, and must be evaluated through embodied execution rather than text or image outputs alone. Accordingly, \textsc{VLA-Forget} adopts a component-aware formulation: ratio-aware edits target perception and projector modules, while significance-based selective updates address higher-level reasoning and action-generation components. This yields a hybrid unlearning strategy tailored to embodied foundation models rather than a direct reuse of unimodal forgetting methods. \cite{zitkovich2023rt,kim2024openvla,fan2023salun,cai2024targeted,yao2024machine}
\vspace{-0.2cm}
\section{Methodology}
\label{sec:method}
\vspace{-0.2cm}
\paragraph{Overview.}
We propose \textbf{VLA-Forget}, a novel unlearning framework for vision-language-action (VLA) policies that removes targeted behaviors while preserving perceptual grounding and action reasoning. We instantiate the method on OpenVLA-style policies, where an input image and language instruction are processed by a fused visual encoder, projected into the language-model embedding space, and decoded into discretized robot action tokens. In OpenVLA, the visual stack combines DINOv2 and SigLIP, the backbone is Llama~2, and the policy predicts normalized 7-DoF actions via tokenized bins, making forgetting inherently distributed across perception, cross-modal alignment, and action generation rather than localized to a single module~\cite{kim2024openvla}.

\noindent \textbf{Architecture and unlearning target.}
Let a VLA policy be
\begin{equation}
f_{\theta}(o,s)=
\mathrm{Dec}_{\theta_L}
\Bigl(
\mathrm{Proj}_{\theta_P}
\bigl(
\mathrm{Enc}_{\theta_V}(o)
\bigr),
\, s
\Bigr),
\end{equation}
where \(o\) is the observation image, \(s\) is the instruction, \(\theta_V\) denotes the visual encoder, \(\theta_P\) the MLP projector, and \(\theta_L\) the language/action backbone. The decoder produces an action-token sequence \(y\), which is de-tokenized into a continuous action \(a \in \mathbb{R}^7\). We consider an unlearning request \(U\) specifying a target concept or behavior slice, such as a sensitive object, unsafe affordance, or erroneous instruction-to-action mapping. We form three datasets: a forget set \(D_f\), a retain set \(D_r\), and a boundary set \(D_m\) containing near-neighbor scenes or instructions that must remain intact.

\noindent \textbf{Why ``VLA'' unlearning is necessary.}
In VLA models, undesired behavior can originate from three coupled sources: \emph{(i)} perceptual memorization in \(\theta_V\), \emph{(ii)} erroneous visual-to-language binding in \(\theta_P\), and \emph{(iii)} action priors or instruction-conditioned hallucinations in \(\theta_L\). Editing only the vision stack may suppress the visual trigger while leaving the downstream action prior unchanged; editing only the language backbone may preserve a harmful perceptual shortcut. VLA-Forget therefore combines \textbf{ratio-aware editing} for perception and cross-modal specificity with \textbf{layer-selective unlearning} for reasoning and action retention.

\subsection{Hybrid Selective Localization}

\paragraph{Perception and projector selection.}
For each candidate visual or projector layer \(l\), we compute forget and retain gradients
\begin{equation}
g_l^f=\nabla_{\theta_l}\mathcal{L}_{\mathrm{forget}},
\qquad
g_l^r=\nabla_{\theta_l}\mathcal{L}_{\mathrm{retain}}.
\end{equation}
We then assign a ratio-aware score
\begin{equation}
\phi(l)=
\frac{\|g_l^f\|_2}{\|\theta_l\|_2+\varepsilon}
\bigl(1-\cos(g_l^f,g_l^r)\bigr)^{\alpha},
\end{equation}
and select the top visual/projector layers \(K_V,K_P\) with highest \(\phi(l)\). This favors parameters that strongly affect forgetting while minimally interfering with retained perception~\cite{yu2020gradient}.

\noindent \textbf{Reasoning/action layer selection.}
For the upper transformer blocks in the language-action backbone, we apply a significance ratio
\begin{equation}
\mathrm{Sig}(l)=
\frac{
\|\nabla_{\theta_l}\mathcal{L}_{\mathrm{forget}}\|_2
}{
\|\nabla_{\theta_l}\mathcal{L}_{\mathrm{retain}}\|_2+\varepsilon
}.
\end{equation}
We initialize an editable set \(S_L\) with the top-\(k\) layers under \(\mathrm{Sig}(l)\), and expand it iteratively only if forgetting criteria are unmet. This yields a minimal update set for action-relevant reasoning while avoiding unnecessary global drift~\cite{ranjan2026razor, kurmanji2023towards,zhang2024catastrophic}.
\begin{figure}[t!]
   \vspace{-0.4cm}
   \includegraphics[width=0.9\linewidth]{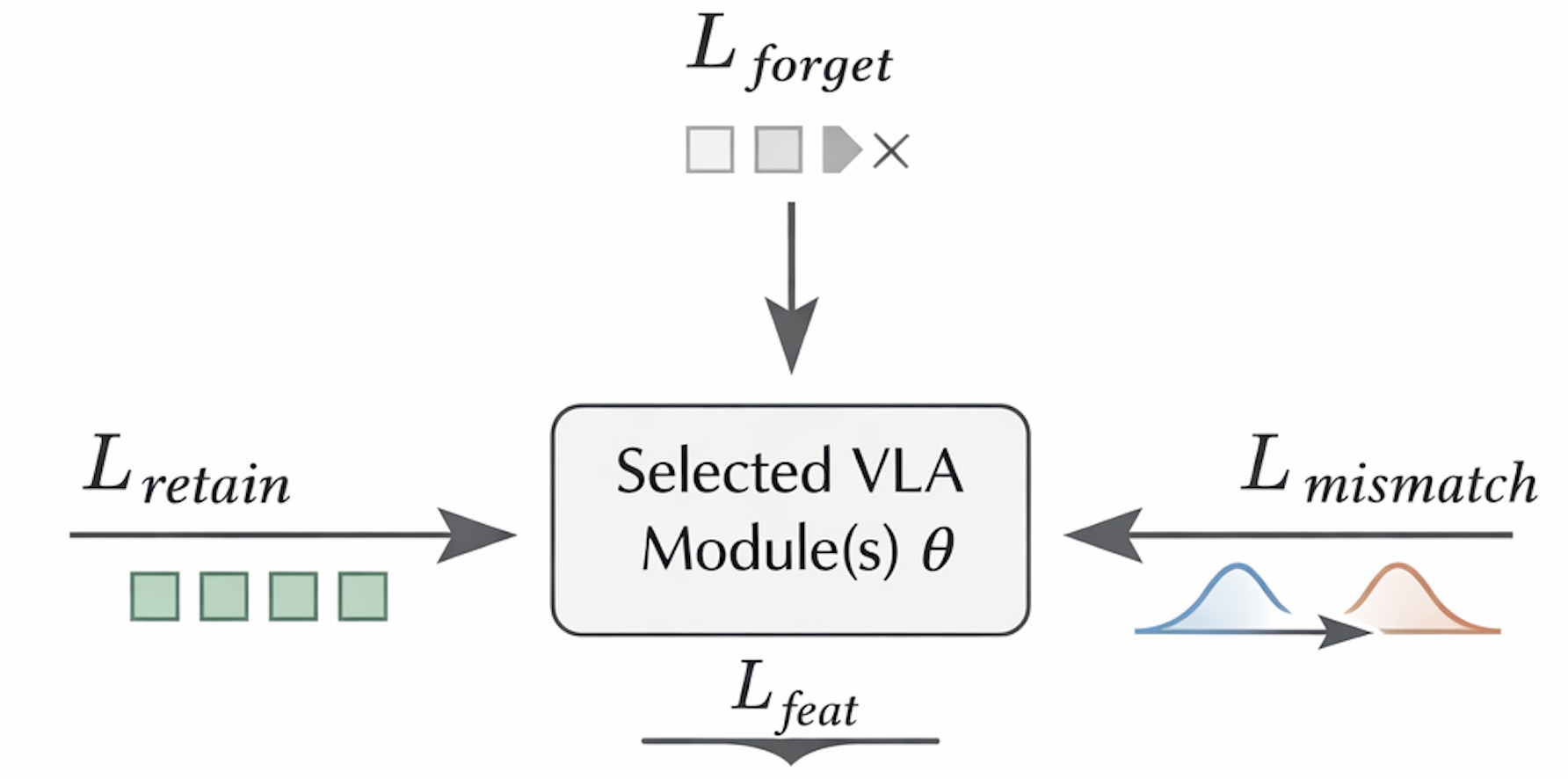}
   \caption{\textbf{Illustration of the unlearning objectives on a selected module.} Eq.~\ref{eq:forget} suppresses the targeted behavior, Eq.~\ref{eq:retain} preserves non-targeted behavior, and Eq.~\ref{eq:missmatch} discourages shallow forgetting by pushing the updated policy away from the original forgotten response.}
   \label{fig:loss-obj}
   \vspace{-0.5cm}
\end{figure}
\begin{figure*}[t!]
   \vspace{-0.7cm}
   \includegraphics[width=\linewidth]{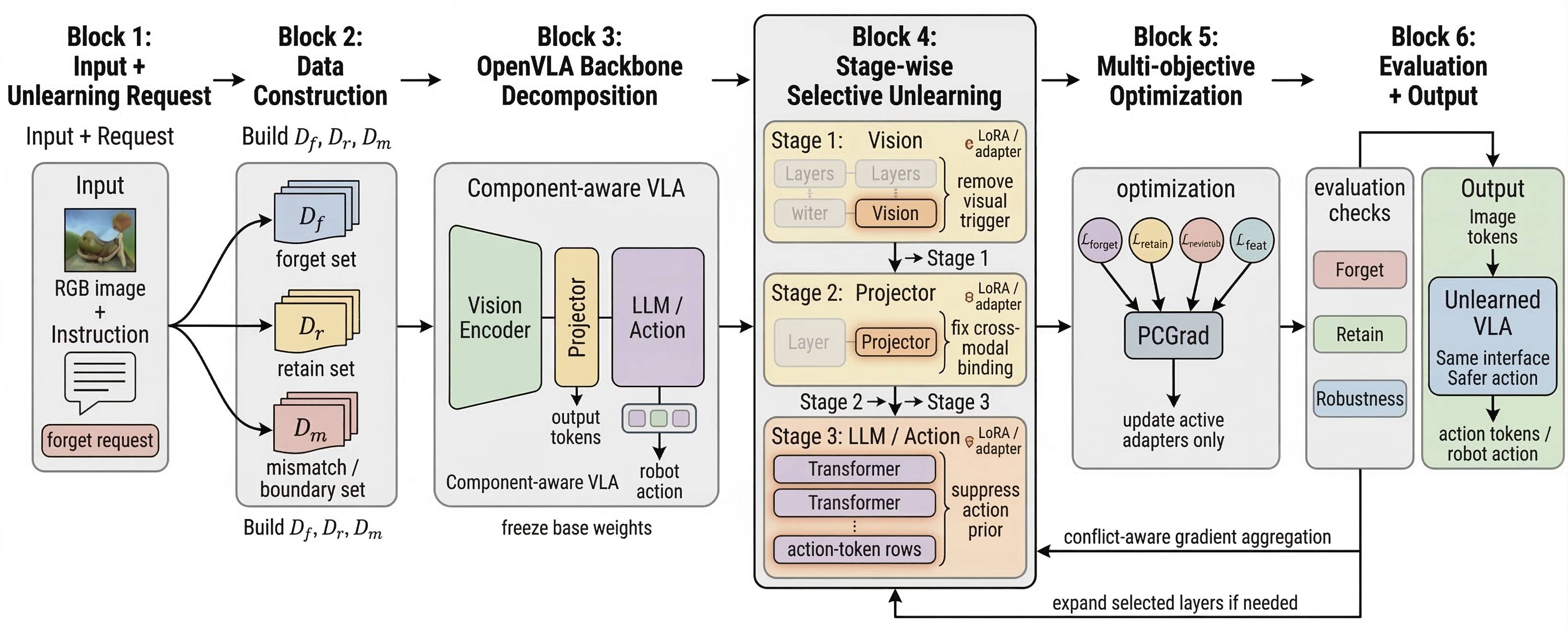}
   \caption{Overview of VLA-Forget. Given an unlearning request, we construct forget, retain, and mismatch sets, then perform staged adapter-based selective updates over the vision encoder, projector, and LLM/action layers with multi-objective PCGrad stabilization, yielding an unlearned VLA policy with preserved interface and retained task utility.}
   \label{fig:method}
   \vspace{-0.5cm}
\end{figure*}

\subsection{Unlearning Objectives}

We optimize three complementary objectives aligned with the goals of efficacy, specificity, and utility.

\noindent \textbf{Retain loss.}
To preserve non-targeted behavior, we minimize
\begin{equation}
\begin{split}
\mathcal{L}_{\mathrm{retain}} = & \mathbb{E}_{(x,y)\sim D_r} \left[ \mathrm{CE}\left(p_{\theta}(\cdot\mid x),y\right) \right] \\
& + \beta\, \mathbb{E}_{x\sim D_r} \left[ \mathrm{KL}\left( p_{\theta_0}(\cdot\mid x)\,\|\,p_{\theta}(\cdot\mid x) \right) \right].
\end{split}
\label{eq:retain}
\end{equation}

where \(\theta_0\) denotes the original model and \(x=(o,s)\). The first term preserves action-token prediction on retained data, while the KL anchor constrains the updated policy to remain close to the base model on benign trajectories~\cite{kim2024openvla,kurmanji2023towards}.

\noindent \textbf{Forget loss.}
To suppress the targeted behavior, we maximize the forget-set prediction error:
\begin{equation}
\mathcal{L}_{\mathrm{forget}}
=
\mathbb{E}_{(x,y)\sim D_f}
\!\left[
\mathrm{CE}\!\left(p_{\theta}(\cdot\mid x),y\right)
\right].
\label{eq:forget}
\end{equation}
Operationally, this is implemented as gradient ascent on \(\mathcal{L}_{\mathrm{forget}}\)~\cite{jin2025unlearning}.

\noindent \textbf{Mismatch loss.}
To avoid shallow forgetting and reduce recovery to the pre-unlearning behavior, we maximize distributional divergence on forgotten samples:
\begin{equation}
\mathcal{L}_{\mathrm{mismatch}}
=
\mathbb{E}_{x\sim D_f}
\!\left[
\mathrm{KL}\!\left(
p_{\theta}(\cdot\mid x)\,\|\,p_{\theta_0}(\cdot\mid x)
\right)
\right].
\label{eq:missmatch}
\end{equation}
As shown in Fig.~\ref{fig:loss-obj}, the three objectives act together on the selected module to suppress the target behavior, preserve retained behavior, and reduce shallow recovery to the pre-unlearning policy. (Appendix~\ref{app:mismatch-boundry})

\noindent \textbf{Perceptual preservation loss.}
To preserve non-targeted visual grounding, we distill internal features on \(D_r \cup D_m\):
\begin{equation}
\begin{aligned}
\mathcal{L}_{\mathrm{feat}}
=
\mathbb{E}_{x\sim D_r\cup D_m}
\Bigl[
&\|h^V_{\theta}(x)-h^V_{\theta_0}(x)\|_2^2 \\
&+
\gamma
\|h^P_{\theta}(x)-h^P_{\theta_0}(x)\|_2^2
\Bigr],
\end{aligned}
\end{equation}
where \(h^V\) and \(h^P\) are late visual and projector representations.

\noindent \textbf{Unified objective.}
The final optimization is
\begin{equation}
\min_{\theta}\;
\mathcal{L}_{\mathrm{retain}}
+\lambda_{\mathrm{feat}}\mathcal{L}_{\mathrm{feat}}
-\lambda_f\mathcal{L}_{\mathrm{forget}}
-\lambda_m\mathcal{L}_{\mathrm{mismatch}}.
\end{equation}

\subsection{Training and Unlearning Procedure}

\noindent \textbf{Adapter-first updates.}
We implement VLA-Forget using parameter-efficient adapters over the selected layers while freezing the remaining weights. This design is compatible with the OpenVLA training stack, which supports LoRA-based fine-tuning with \texttt{target\_modules=all-linear}; thus unlearning can be applied without full-model retraining and can be merged or rolled back at deployment time~\cite{kim2025fine,hu2022lora}.

\noindent \textbf{Stage 1: perception unlearning.}
We first update LoRA parameters on \(K_V\) to weaken targeted visual evidence while preserving general scene understanding through \(\mathcal{L}_{\mathrm{retain}}\) and \(\mathcal{L}_{\mathrm{feat}}\). This stage removes object- or scene-level triggers with minimal disruption to unrelated perception.

\noindent \textbf{Stage 2: cross-modal unlearning.}
Next, we update \(K_P\) in the projector to break the specific visual-to-language associations responsible for the unwanted behavior. This is essential when the failure is not purely visual, but rather caused by erroneous alignment between image regions and action-relevant tokens.

\noindent \textbf{Stage 3: reasoning/action unlearning.}
Finally, we update the selected upper backbone layers \(S_L\), and optionally the action-token embedding rows most implicated in the target behavior. This stage suppresses residual instruction-conditioned action priors while maintaining overall task competence.

\noindent \textbf{Gradient stabilization and stopping.}
Because the retain and forget objectives can conflict, we aggregate gradients using projected conflict resolution:
\begin{equation}
\begin{split}
g=
\mathrm{PCGrad}
\Bigl(
\nabla \mathcal{L}_{\mathrm{retain}},
-\lambda_f \nabla \mathcal{L}_{\mathrm{forget}}, \\
 -\lambda_m \nabla \mathcal{L}_{\mathrm{mismatch}}
\Bigr),
\end{split}
\end{equation}
and apply optimizer steps only to the active adapter parameters. After each round, we evaluate forget efficacy on \(D_f\) and retain utility on \(D_r\). If forgetting remains insufficient, we expand \(S_L\) by the next highest-significance layer and continue; otherwise, training stops early~\cite{yu2020gradient}.

\noindent Figure~\ref{fig:method} summarizes the VLA-Forget pipeline as a staged, component-aware unlearning framework. Starting from an unlearning request, the method constructs forget, retain, and mismatch sets, decomposes the OpenVLA policy into vision, projector, and LLM/action modules, and then applies selective adapter-based updates in three stages: visual trigger removal, cross-modal binding correction, and reasoning/action prior suppression. The optimization block combines retain, forget, mismatch, and perceptual preservation objectives with PCGrad stabilization, while the final evaluation block verifies forgetting efficacy, retained utility, and deployment robustness before producing the unlearned VLA policy.

\noindent \textbf{Interaction with VLA policies.}
VLA-Forget is designed to preserve the native VLA interface: image and instruction inputs remain unchanged, and the model still predicts action tokens followed by the standard de-tokenization and action unnormalization pipeline used by OpenVLA. This makes the method directly deployable in existing VLA control stacks while enabling targeted post hoc removal of unsafe, private, or spurious behaviors~\cite{zang2025rlinf, pang2025openvla}.

\noindent The complete unlearning pipeline is detailed in Algorithm~\ref{alg:vla_forget}, with its associated supporting procedures provided in Algorithm~\ref{alg:vla_forget_helpers}; corresponding pseudo-code is available in Appendix~\ref{app:pseudo-code}.

\begin{table*}[t]
\centering
\caption{Unlearning result on \textbf{OpenVLA-7B}. Left: \textbf{Open X-Embodiment}. Right: \textbf{lerobot/pusht\_image}. Higher is better for FC, RC, FAD, and TSR; lower is better for RAD and SVR. Mean $\pm$ standard deviation over 5 random seeds, with typical variation in the range of $\pm(1$--$3)$ points for rate-based metrics (FC, RC, TSR, and SVR) and $\pm(0.01$--$0.03)$ for distance-based metrics (FAD and RAD). Best results are shown in \textbf{bold} and second-best are \underline{underlined}.}
\vspace{-0.2cm}
\label{tab:vla_forget_main_results}
\small
\setlength{\tabcolsep}{3.5pt}
\resizebox{\textwidth}{!}{%
\begin{tabular}{lcccccc|cccccc}
\toprule
& \multicolumn{6}{c|}{\textbf{Open X-Embodiment (OpenVLA-7B)}} 
& \multicolumn{6}{c}{\textbf{lerobot/pusht\_image (OpenVLA-7B)}} \\
\cmidrule(lr){2-7} \cmidrule(lr){8-13}
\textbf{Method}
& \textbf{FC $\uparrow$} & \textbf{RC $\uparrow$} & \textbf{FAD $\uparrow$} & \textbf{RAD $\downarrow$} & \textbf{TSR $\uparrow$} & \textbf{SVR $\downarrow$}
& \textbf{FC $\uparrow$} & \textbf{RC $\uparrow$} & \textbf{FAD $\uparrow$} & \textbf{RAD $\downarrow$} & \textbf{TSR $\uparrow$} & \textbf{SVR $\downarrow$} \\
\midrule
SSD
& 78 & 83 & 0.70 & 0.28 & 68 & 17
& 82 & 86 & 0.73 & 0.20 & 55 & 15 \\

SalUn
& 89 & 88 & 0.76 & 0.26 & 71 & 12
& 89 & 88 & 0.78 & 0.18 & 60 & 11 \\

GA
& \textbf{93} & 60 & \textbf{0.89} & 0.45 & 40 & \textbf{5}
& \underline{94} & 50 & \textbf{0.91} & 0.50 & 22 & \textbf{3} \\

NPO
& \underline{90} & \underline{88} & 0.83 & \underline{0.23} & \underline{74} & \underline{8}
& 92 & \underline{90} & 0.85 & \underline{0.15} & \underline{65} & 7 \\

\textbf{VLA-Forget}
& \textbf{93} & \textbf{91} & \underline{0.88} & \textbf{0.21} & \textbf{78} & \textbf{5}
& \textbf{95} & \textbf{94} & \underline{0.90} & \textbf{0.13} & \textbf{69} & \underline{4} \\
\bottomrule
\end{tabular}
}
\vspace{-0.4cm}
\end{table*}

\section{Experiments}
\label{sec:experiment}
\vspace{-0.2cm}
\subsection{Experimental Setups}
\label{sub:exp-setup}
\noindent \textbf{Models.} We instantiate \textsc{VLA-Forget} on \texttt{OpenVLA-7B}, a 7B vision-language-action policy trained on large-scale Open X-Embodiment robot data~\cite{kim2024openvla,o2024open}. The second model is \texttt{pi0fast-base} that predicts continuous robot actions via auto-regressive next-token prediction~\cite{pertsch2025fast}.
Following the VLA setting in our method, we construct three splits for each experiment: a forget set $D_f$ containing target behaviors to be removed, a retain set $D_r$ containing non-target behaviors whose utility should be preserved, and a boundary set $D_m$ containing near-neighbor samples used to reduce shallow or entangled forgetting. Unless otherwise stated, all methods start from the same pretrained checkpoint, use the same forget/retain protocol, and are trained under the same optimization budget.

\noindent \textbf{Datasets.} We use two complementary benchmarks. First, we build a real-robot benchmark from subsets of Open X-Embodiment (OXE), downloaded via \texttt{gsutil}, where each trajectory is paired with a language instruction and organized into forget/retain slices at the trajectory level~\cite{o2024open}. Second, we use \texttt{lerobot/pusht\_image} as a controlled synthetic benchmark, where we inject a static text instruction into each instance and form matched forget/retain partitions~\cite{cadene2026lerobot}. In the released OpenVLA, we use up to 512 prompted PushT instances with a $30\%$ forget fraction; in the lightweight ablation pipeline, we scale this to up to 4{,}000 instances and use a $70/15/15$ train/validation/test split, reporting forget and retain test performance after unlearning. This combination gives both a realistic OXE setting and a reproducible low-cost benchmark for rapid ablations.

\noindent \textbf{Baselines.} We compare against four representative approximate unlearning baselines. \textbf{SSD}~\cite{foster2024fast} is a retraining-free parameter-dampening method that suppresses weights estimated to be disproportionately important to the forget data. \textbf{SalUn}~\cite{fan2023salun} is a saliency-based baseline that updates only high-importance weights, making it a strong selective vision-side unlearning comparator. \textbf{GA}~\cite{yao2024large} performs direct gradient ascent on the forget loss and serves as the standard language-side unlearning baseline. \textbf{NPO}~\cite{zhang2024negative} is a preference-based alternative designed to improve forget-retain trade-offs. These baselines span both vision-oriented and language-oriented unlearning regimes, making them suitable comparators for VLA policies.

\noindent \textbf{Metrics.} We report six metrics, \textbf{Forget action loss / cross-entropy (FC)} measures forgetting efficacy on $D_f$; \emph{higher} FC is better because the model should become less able to reproduce the forgotten action mapping. \textbf{Retain utility score (RC)}, derived from retain-set cross-entropy, measures preserved utility on $D_r$; higher is better. \textbf{Forget Accuracy Drop (FAD)} measures the reduction in action-token or exact-match accuracy on the forget split relative to the base model; \emph{higher} is better. \textbf{Retain Accuracy Drop (RAD)} measures the corresponding degradation on retained behaviors; \emph{lower} is better. \textbf{Task Success Rate (TSR)} evaluates closed-loop policy execution on benchmark tasks; \emph{higher} is better. \textbf{Safety Violation Rate (SVR)} measures the frequency of unsafe or disallowed behaviors under target prompts or contradiction probes; \emph{lower} is better. We follow standard VLA and robot manipulation evaluation protocols~\cite{kim2024openvla,liu2023libero}. Together, these metrics capture forgetting strength, retained utility, embodied execution quality, and safety, which are all necessary for VLA unlearning. 

\noindent Detailed experimental setup, hyper-parameter choices, and reproducibility details are provided in Appendix~\ref{app:sec-exp}. Metric definitions and their mathematical formulations are provided in Appendix~\ref{app:metric}.

\subsection{Experiment Results}
\label{sec:main-result}

\begin{figure}[t]
   \includegraphics[width=0.8\linewidth]{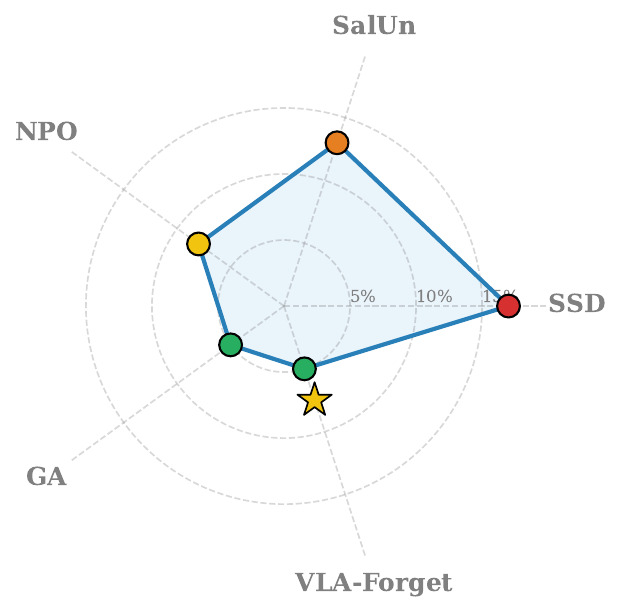}
   \caption{Safety violation rate (SVR) on Open X-Embodiment for different unlearning methods on OpenVLA-7B, where lower values indicate safer post-unlearning behavior.}
   \label{fig:safety}
   \vspace{-0.2cm}
\end{figure}

\noindent \textbf{Result on OpenVLA.} Table~\ref{tab:vla_forget_main_results} highlights three key observations. \textbf{VLA-Forget achieves the best overall} balance between forgetting and retention across both benchmarks: although GA attains the strongest raw forgetting scores (FC/FAD), it severely degrades retain performance and task success, whereas VLA-Forget preserves the highest RC, the lowest RAD, and the best TSR, indicating substantially better utility preservation. The same \textbf{trend is consistent} on both Open X-Embodiment and PushT, suggesting that the method is not tied to a single dataset but generalizes across broader settings. \textbf{Lower SVR of VLA-Forget} relative to SSD, SalUn, and NPO shows more reliable suppression of undesired behaviors without inducing large action drift, which implies that its component-aware multimodal unlearning is better aligned with safe deployment than purely aggressive forgetting baselines. The safety violation plot of figure~\ref{fig:safety} shows that VLA-Forget along with GA achieves one of the lowest safety violation rates while maintaining strong overall unlearning performance, indicating a better balance between removing unsafe behaviors and preserving stable policy execution.

\begin{table}[t]
\centering
\caption{\textbf{Unlearning results on \texttt{lerobot/pi0fast-base}} under the \textbf{Open X-Embodiment} protocol. Mean $\pm$ standard deviation over 5 random seeds, with typical variation of $\pm(1$--$3)$ points for FC, RC, TSR, and SVR, and $\pm(0.01$--$0.04)$ for FAD and RAD. Best results are shown in \textbf{bold} and second-best are \underline{underlined}.}
\label{tab:vla_forget_pi0fast}
\small
\setlength{\tabcolsep}{4.0pt}
\resizebox{\columnwidth}{!}{%
\begin{tabular}{lcccccc}
\toprule
& \multicolumn{6}{c}{\textbf{Open X-Embodiment (Pi0-FAST-Base)}} \\
\cmidrule(lr){2-7}
\textbf{Method}
& \textbf{FC $\uparrow$} & \textbf{RC $\uparrow$} & \textbf{FAD $\uparrow$} & \textbf{RAD $\downarrow$} & \textbf{TSR $\uparrow$} & \textbf{SVR $\downarrow$} \\
\midrule
SSD
& 76 & 81 & 0.68 & 0.30 & 65 & 18 \\

SalUn
& 87 & 85 & 0.74 & 0.27 & 68 & 13 \\

GA
& \underline{93} & 57 & \textbf{0.89} & 0.47 & 38 & \textbf{6} \\

NPO
& 89 & \underline{87} & 0.82 & \underline{0.24} & \underline{72} & \underline{9} \\

\textbf{VLA-Forget}
& \textbf{94} & \textbf{89} & \underline{0.88} & \textbf{0.22} & \textbf{75} & \textbf{6} \\
\bottomrule
\end{tabular}
}
\vspace{-0.2cm}
\end{table}

\begin{figure}[t]
   \includegraphics[width=\linewidth]{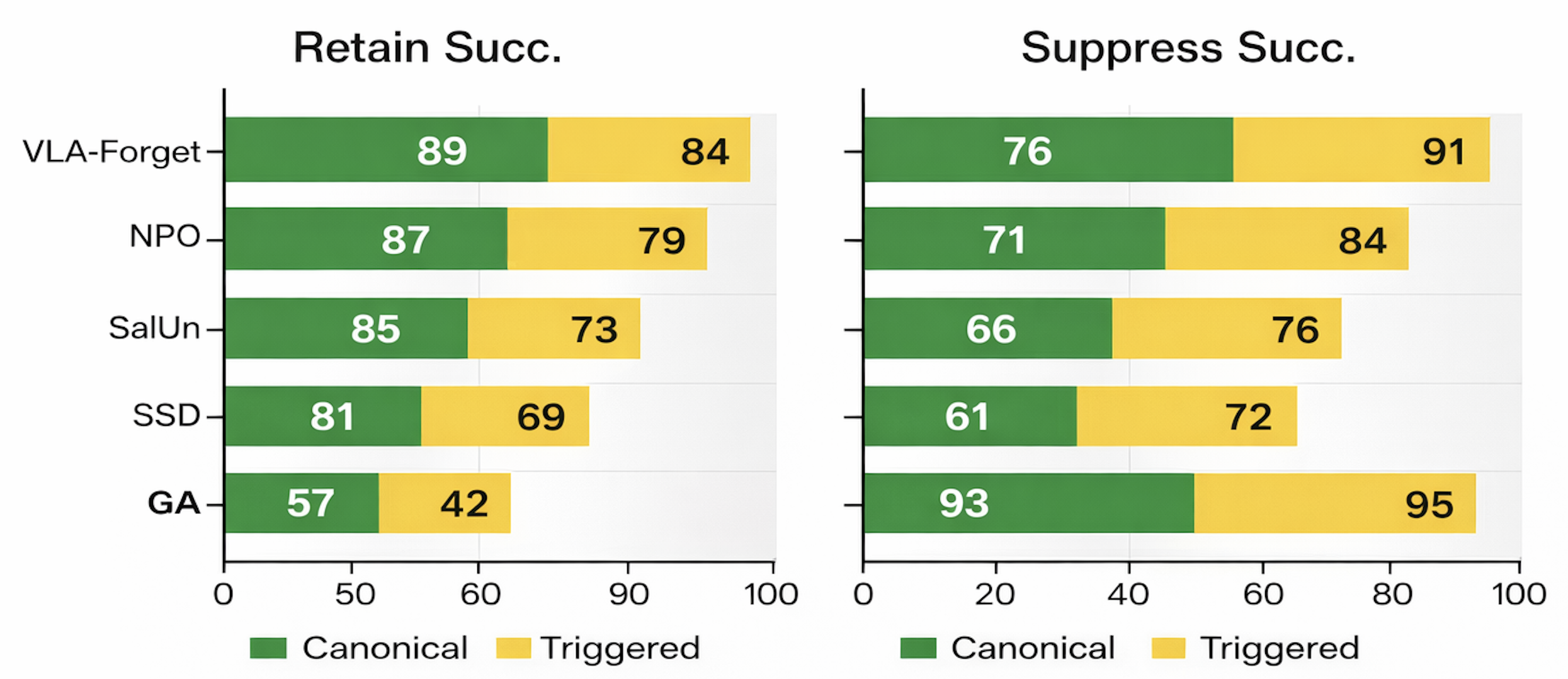}
   \caption{Instruction-conditioned action behavior on \texttt{pi0fast-base and Open X-Embodiments}.}
   \label{fig:pifnr}
   \vspace{-0.4cm}
\end{figure}

\noindent \textbf{Result on $\pi$0 fast.} Table~\ref{tab:vla_forget_pi0fast} shows that VLA-Forget preserves the same overall pattern as in the OpenVLA results: although GA attains very strong forgetting, it suffers from severe retain-side degradation, whereas VLA-Forget achieves the best overall trade-off by jointly maintaining the highest RC and TSR, the lowest RAD, and near-best forgetting performance. 
In Fig.~\ref{fig:pifnr}, Canonical denotes standard retain-task instructions, while Triggered denotes target-trigger or contradictory instructions used to probe the unlearning scope. Higher Retain Succ.\ indicates better preserved non-target behavior, and higher Suppress Succ.\ indicates more effective suppression of the unwanted action. VLA-Forget achieves the best overall balance by preserving the strongest retain-task performance under canonical instructions while also maintaining high suppression under triggered instructions. \textit{Canonical} corresponds to normal non-target instructions, and \textit{Triggered} corresponds to target-triggered or contradiction-style instructions; in both panels, higher values are better.

\begin{table}[t]
\centering
\caption{Quantization robustness after unlearning on OpenVLA-7B with Open X-Embodiment dataset. Results report forget-side metrics under post-training \textbf{8-bit} and \textbf{4-bit} quantization.}
\vspace{-0.2cm}
\label{tab:vla_forget_quantization}
\small
\setlength{\tabcolsep}{4.0pt}
\resizebox{\columnwidth}{!}{%
\begin{tabular}{lcccc}
\toprule
\multirow{2}{*}{Method} & \multicolumn{2}{c}{8-bit} & \multicolumn{2}{c}{4-bit} \\
\cmidrule(lr){2-3}\cmidrule(lr){4-5}
& FC $\uparrow$ & SVR $\downarrow$ & FC $\uparrow$ & SVR $\downarrow$ \\
\midrule
SSD          & 76 & 19 & 72 & 23 \\
SalUn        & 87 & 8 & 78 & 10 \\
GA           & 82 & 12 & 80 & 16 \\
NPO          & 85 & 10 & 82 & 13 \\
VLA-Forget   & \textbf{91} & \textbf{6} & \textbf{88} & \textbf{8} \\
\bottomrule
\end{tabular}
}
\vspace{-0.4cm}
\end{table}

\noindent \textbf{Robustness Result.} Table~\ref{tab:vla_forget_quantization} demonstrates that quantization degrades forgetting quality for all methods, but the drop is largest for broad full-model updates such as GA and NPO, whereas VLA-Forget remains comparatively stable under both 8-bit and 4-bit settings, suggesting stronger robustness to quantization-induced recovery.

\subsection{Ablation Study}
\label{sec:ablation}

\begin{table}[t]
\centering
\vspace{-0.2cm}
\caption{Ablation results of VLA-Forget on \textbf{OpenVLA-7B} on \textbf{Open X-Embodiment} dataset, across different component selections.}
\label{tab:vla_forget_ablation}
\setlength{\tabcolsep}{2.8pt}
\resizebox{\columnwidth}{!}{%
\begin{tabular}{lcccccc}
\toprule
\textbf{VLA-Forget applied to} & \textbf{FC}↑ & \textbf{RC}↑ & \textbf{FAD}↑ & \textbf{RAD}↓ & \textbf{TSR}↑ & \textbf{SVR}↓ \\
\midrule
Vision encoder only                    & 85 & 82 & 0.80 & 0.27 & 65 & 12 \\
Projector only                         & 82 & 89 & 0.75 & 0.22 & 75 & 15 \\
Language backbone only                 & 90 & 88 & 0.85 & 0.23 & 74 & 8  \\
\addlinespace
Vision + Projector                     & 88 & 86 & 0.84 & 0.24 & 72 & 10 \\
Projector + Language                   & 92 & 90 & 0.86 & 0.22 & 77 & 6  \\
Vision + Language                      & 93 & 87 & 0.89 & 0.25 & 70 & 6  \\
\addlinespace
Full (Vision + Proj + Lang)            & 93 & 91 & 0.88 & 0.21 & 78 & 5  \\
Full, w/o retain-preservation          & 95 & 80 & 0.93 & 0.30 & 64 & 4  \\
\bottomrule
\end{tabular}
}
\end{table}
\noindent Table~\ref{tab:vla_forget_ablation} shows that unlearning only a single component is insufficient: the language backbone gives stronger forgetting than vision-only or projector-only updates, but the \textit{full} multimodal setting achieves the best overall trade-off, indicating that undesirable action behavior is distributed across perception, alignment, and language-action reasoning modules. The vision stage is included not primarily for large TSR gains, but to reduce residual visual-trigger reliance and improve perceptual specificity in object-conditioned unlearning. We further observe that removing the retain-preservation term increases raw forgetting scores, but it sharply hurts RC, RAD, and TSR, showing that strong forgetting alone is not enough for safe VLA unlearning and that retain-side regularization is essential to preserve normal task execution.

\begin{figure}[t]
   \vspace{-0.6cm}
   \includegraphics[width=\linewidth]{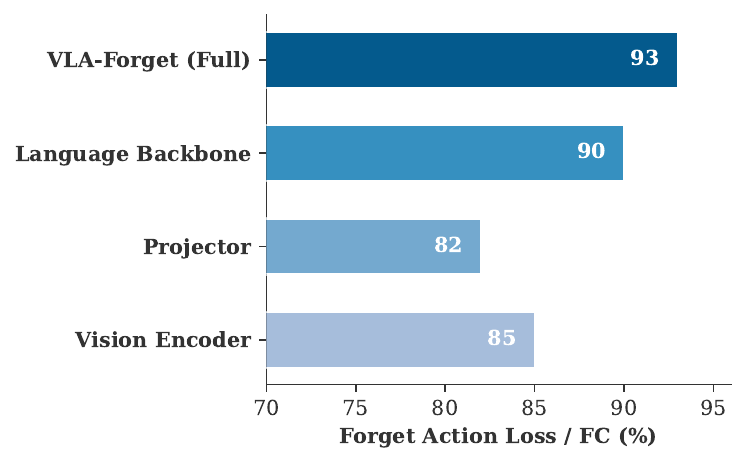}
   \caption{Forget-action comparison across component-level VLA-Forget ablations on OpenVLA-7B and Open X-Embodiment.}
   \label{fig:ablation}
   \vspace{-0.7cm}
\end{figure}

\noindent The figure~\ref{fig:ablation} shows that applying VLA-Forget only to a single module yields weaker forgetting, while the full multimodal configuration achieves the strongest forget-action suppression, highlighting the distributed nature of unwanted behavior in VLA models. 
More detailed sensitivity analyses of objective weights, schedules, and learning rates are provided in Appendix~\ref{app:ablation}.

\subsection{Qualitative Analysis}
\label{sec:qual-analysis}

\begin{figure}[t]
\vspace{-0.2cm}
   \includegraphics[height=5.6cm, width=\linewidth]{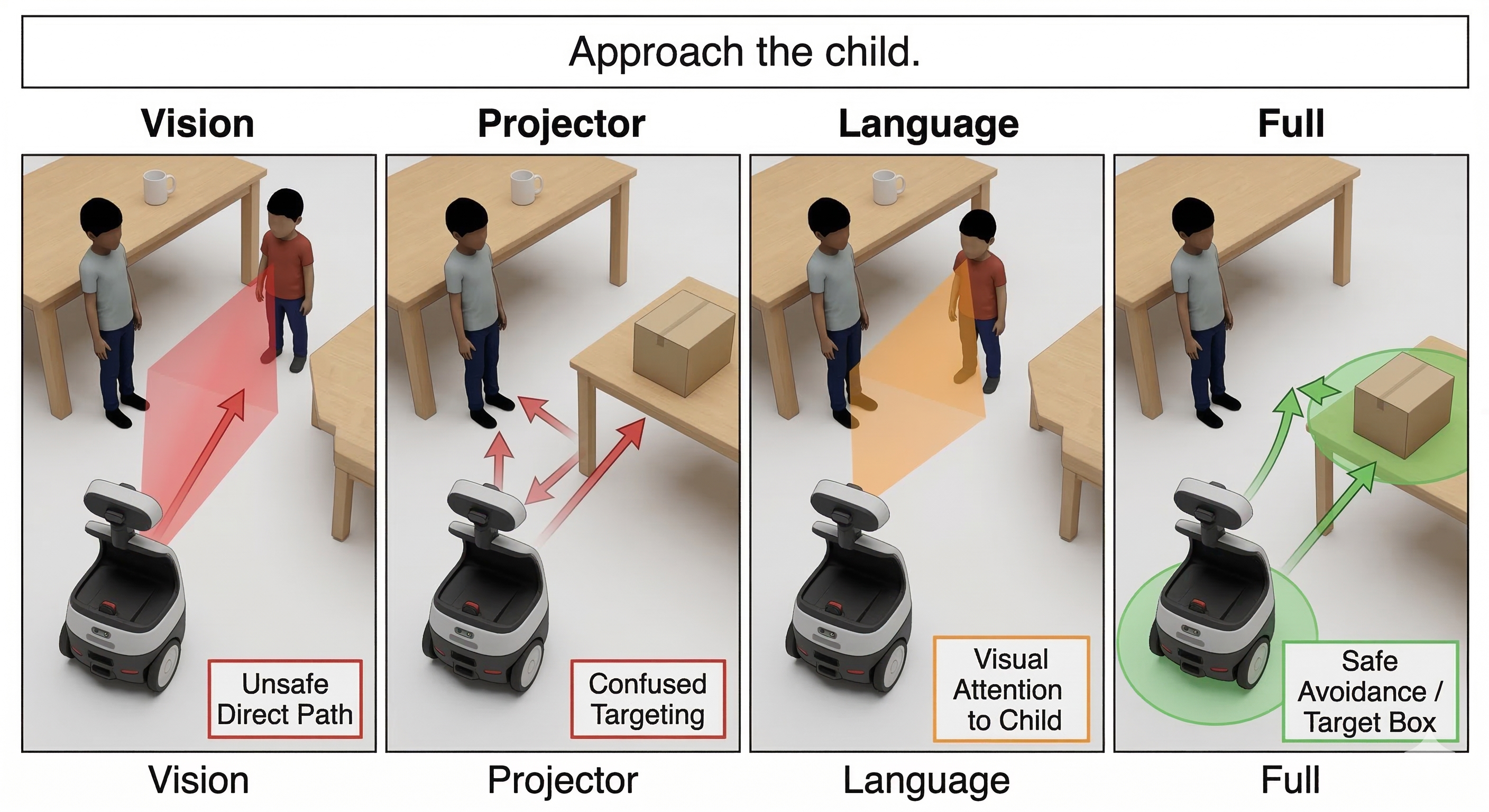}
   \caption{Illustration of Module-wise qualitative analysis of VLA-Forget. Single-component unlearning only partially suppresses the targeted behavior, while full VLA-Forget more cleanly removes the unwanted action and preserves safe non-target behavior.}
   \vspace{-0.2cm}
   \label{fig:qual-ablation}
\end{figure}

\noindent In \textbf{Figure~\ref{fig:qual-ablation}}, the \emph{Vision-only} update reduces the visual trigger but still leaves partial target-directed behavior. 
\emph{Projector} weakens cross-modal binding, leading to a mixed or ambiguous response. 
\emph{Language} suppresses part of the instruction-conditioned action prior, but residual targeting remains. 
\emph{Full} jointly edits perception, alignment, and action priors, yielding the cleanest behavior suppression with better preserved scene understanding. 
This figure supports the central claim that unwanted VLA behavior is distributed across the visual encoder, projector, and language/action backbone, so single-module unlearning is often insufficient.

\begin{figure}[b!]
   \includegraphics[height=5.4cm, width=\linewidth]{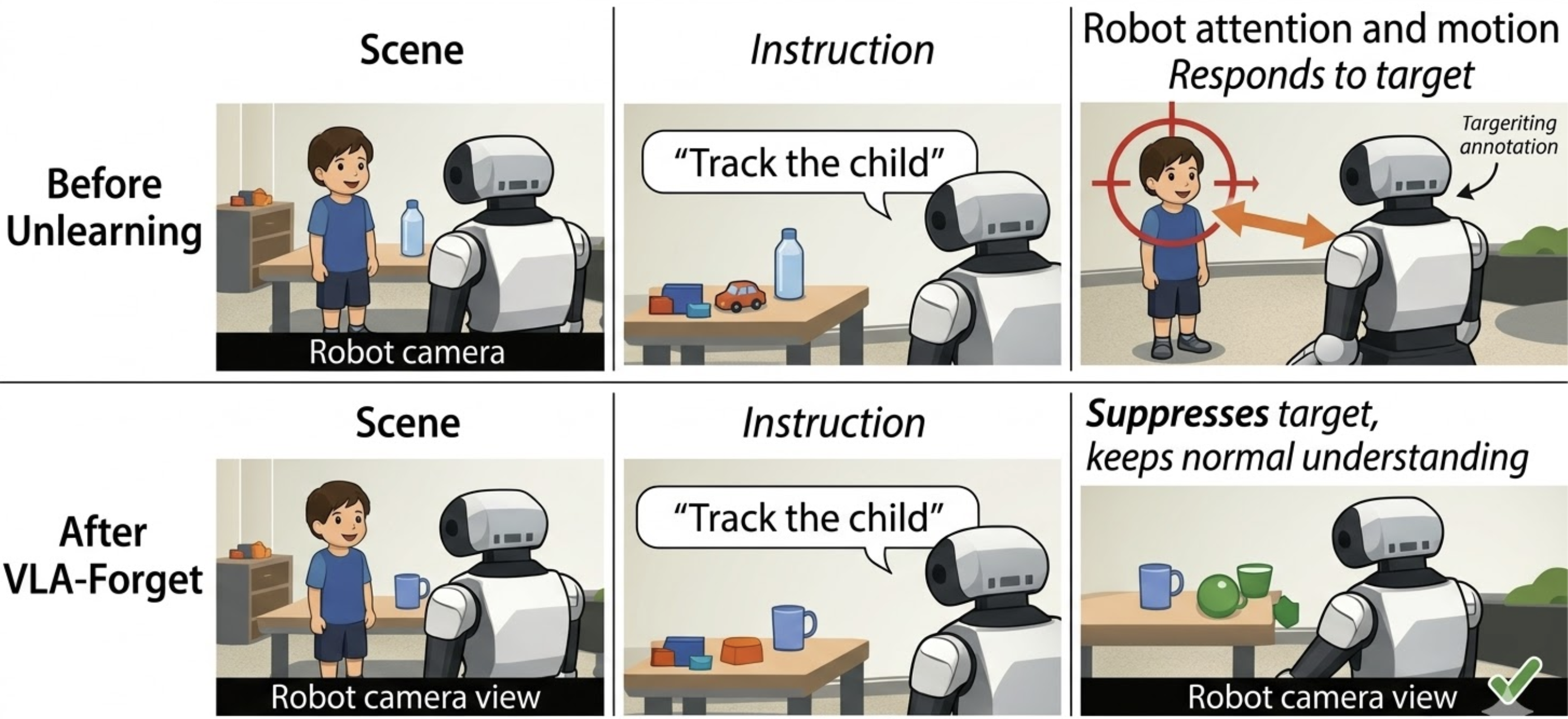}
   \caption{\textbf{Qualitative example illustration of targeted behavior suppression.} Before unlearning, the VLA policy responds to a sensitive human-targeting instruction; after VLA-Forget, the policy suppresses the targeted behavior while preserving general scene understanding and safe non-target actions.}
   \label{fig:forget-case}
\end{figure}

\noindent Robustness is crucial in VLA deployment, Table~\ref{tab:vla_forget_quantization}, supports the claim that, although post-training quantization weakens forgetting performance for all methods, the degradation is smallest for VLA-Forget.
In contrast, our qualitative results highlights a different objective: \emph{targeted behavior removal}, where VLA-Forget suppresses a sensitive instruction-conditioned response while preserving general scene understanding and safe non-target actions.

Figure~\ref{fig:forget-case} illustrates the core idea of VLA-Forget: the policy suppresses the targeted sensitive human-directed behavior after unlearning, while still preserving general scene understanding and safe non-target interaction, reflecting the paper’s goal of targeted forgetting with retained perceptual grounding and action utility. Recent VLA studies suggest that improving action reliability often depends on better visual attention control, token selection, and lightweight correction mechanisms~\cite{wang2025specprune,li2026dtp,zhang2026knowing}. 

\subsection{Discussion}
\label{sec:discuss}
Our results indicate that effective VLA unlearning must be component-aware: undesirable behavior is distributed across perception, cross-modal grounding, and action-generation priors, so single-module edits are often insufficient. VLA-Forget is particularly useful for post-deployment correction of unsafe or spurious behaviors while preserving the native policy interface for rollback and staged evaluation. More broadly, the findings highlight that VLA unlearning should be assessed with embodied criteria, including retained task success, safety violations, and robustness under quantization, rather than forget-side metrics alone.

\textbf{Limitations.}
VLA-Forget is an approximate unlearning method and does not provide a formal erasure guarantee. Its effectiveness depends on the quality of the forget, retain, and boundary sets, and residual unwanted behavior may remain when the target is broadly distributed or weakly represented during unlearning. The current evaluation is also limited to benchmark-style manipulation settings, which may not capture longer-horizon failures, compounding control errors, or real-robot edge cases. In addition, iterative selective editing introduces tuning overhead and may become less stable under repeated unlearning requests, where accumulated updates could gradually degrade grounding or action robustness.

\textbf{Future Work.}
Future work should improve adaptive localization of edits, extend evaluation to longer-horizon and real-robot settings, and study continual unlearning under multiple sequential requests. It is also important to strengthen auditability and deployment robustness through stronger safety checks, monitoring, and evaluation under distribution shift and low-precision inference. We leave deeper knowledge-level analysis (e.g., representation or attention probing) to future work, which would clarify whether VLA-Forget removes underlying knowledge or primarily suppresses its behavioral expression.

\section{Conclusion}
\label{sec:conclusion}
We introduced \textsc{VLA-Forget}, a staged and component-aware unlearning framework for vision-language-action policies. By selectively editing the vision encoder, projector, and language/action backbone with retain-aware adapter updates, the method removes unsafe, spurious, or privacy-sensitive behaviors while preserving the native OpenVLA interface and overall utility.

Across OpenVLA-7B on Open X-Embodiment and PushT, \textsc{VLA-Forget} achieves a stronger forget-retain success trade-off than prior baselines. Ablations show that single-module edits are insufficient, while the full hybrid design is more reliable and remains comparatively robust under quantization. Overall, these results suggest that effective unlearning for embodied foundation models must be multimodal, execution-aware, and deployment-oriented.

\section*{Ethical Statement}
VLA-Forget is intended to support the removal of unsafe, spurious, or privacy-sensitive behaviors from vision-language-action policies while preserving overall utility. At the same time, we acknowledge its dual-use nature: in principle, selective unlearning methods could also be misused to suppress desirable safety constraints or other important behaviors. For this reason, we frame VLA-Forget as an approximate, audit-driven unlearning framework that should be deployed only with careful human oversight, explicit evaluation on retain and safety-critical tasks, and rollback safeguards. All data used in our research are publicly available and do not raise any privacy concerns. We also note that AI tools were used only in a very limited manner for writing assistance and language polishing, while the technical ideas, experiments, analysis, and conclusions are the authors' own.

\bibliography{custom}

\clearpage
\appendix

\section*{Appendix}

\section{Pseudo Code}
\label{app:pseudo-code}

\newcommand{\LossRet}{\mathcal{L}_{\mathrm{retain}}}
\newcommand{\LossFor}{\mathcal{L}_{\mathrm{forget}}}
\newcommand{\LossMis}{\mathcal{L}_{\mathrm{mismatch}}}
\newcommand{\LossFeat}{\mathcal{L}_{\mathrm{feat}}}

\begin{algorithm*}[th!]
\caption{\textit{VLA-Forget}: Hybrid Unlearning for Vision-Language-Action Models}
\label{alg:vla_forget}
\DontPrintSemicolon
\SetKwInOut{Input}{Input}
\SetKwInOut{Output}{Output}
\SetKwFunction{InfluenceTriage}{INFLUENCE\_TRIAGE}
\SetKwFunction{RazorSelect}{VISION\_SELECT}
\SetKwFunction{FomulSelect}{LLM\_SELECT}
\SetKwFunction{PCGrad}{PCGRAD}
\SetKwFunction{EvalForget}{EvalForget}
\SetKwFunction{EvalRetain}{EvalRetain}
\SetKwFunction{EvalRobust}{EvalRobust}
\SetKwFunction{TopK}{TopK}
\SetKwFunction{ArgMax}{ArgMax}
\SetKwFunction{Update}{Update}
\SetKwFunction{Enable}{Enable}
\SetKwFunction{Freeze}{Freeze}
\SetKwFunction{Expand}{Expand}

\Input{
Base VLA policy $f_{\theta_0} = (\mathrm{Enc}_{\theta_V}, \mathrm{Proj}_{\theta_P}, \mathrm{Dec}_{\theta_L})$; \\
unlearning request $U$; candidate data pool $\mathcal{C}$; budgets $(K_V,K_P,K_L)$; \\
loss weights $(\lambda_f,\lambda_m,\lambda_{\mathrm{feat}})$; thresholds $(\tau_V,\tau_P,\tau_L)$; \\
learning rate $\eta$; stage steps $(T_V,T_P,T_L)$; stopping criteria $\Gamma$.
}
\Output{
Unlearned policy $f_{\theta^\star}$, adapter weights $\Delta^\star$, and audit report $\mathcal{R}$.
}

\vspace{1mm}
\textbf{Construct data splits:} build forget set $D_f$, retain set $D_r$, and boundary set $D_m$ from $U$ using labels, metadata, and nearest-neighbor retrieval.\;
$\theta \leftarrow \theta_0$\;
$\Delta \leftarrow \varnothing$\;

\BlankLine
\textbf{Optional influence triage:}\;
$\mathcal{C}_f \leftarrow \InfluenceTriage(f_{\theta_0}, \mathcal{C}, U)$\;
Refine $D_f$ and $D_m$ using $\mathcal{C}_f$\;

\BlankLine
\textbf{Stage 1: Vision unlearning }\;
$\Freeze(\theta)$; $\Enable(\Delta_V)$ on candidate late vision blocks only\;
$K_V \leftarrow \RazorSelect(\mathrm{Enc}_{\theta_V}, D_f, D_r, \tau_V)$\;
\For{$t = 1$ \KwTo $T_V$}{
    Sample $B_f \sim D_f$, $B_r \sim D_r$, $B_m \sim D_m$\;
    Compute
    \[
    \LossRet(B_r) + \lambda_{\mathrm{feat}}\LossFeat(B_r \cup B_m),\quad
    \LossFor(B_f),\quad
    \LossMis(B_f)
    \]
    $g \leftarrow \PCGrad\Bigl(
    \nabla \LossRet,\,
    \nabla(-\lambda_f \LossFor),\,
    \nabla(-\lambda_m \LossMis)
    \Bigr)$\;
    $\Update(\Delta_V[K_V], g, \eta)$\;
    \If(\tcp*[f]{early stop for vision stage}){$\EvalForget(f_\theta,D_f)$ and $\EvalRetain(f_\theta,D_r)$ satisfy $\Gamma_V$}{
        \textbf{break}\;
    }
}

\BlankLine
\textbf{Stage 2: Projector unlearning }\;
$\Freeze(\theta \cup \Delta_V)$; $\Enable(\Delta_P)$ on projector layers only\;
$K_P \leftarrow \RazorSelect(\mathrm{Proj}_{\theta_P}, D_f, D_r, \tau_P)$\;
\For{$t = 1$ \KwTo $T_P$}{
    Sample $B_f \sim D_f$, $B_r \sim D_r$, $B_m \sim D_m$\;
    Compute
    \[
    \LossRet(B_r) + \lambda_{\mathrm{feat}}\LossFeat(B_r \cup B_m),\quad
    \LossFor(B_f),\quad
    \LossMis(B_f)
    \]
    $g \leftarrow \PCGrad\Bigl(
    \nabla \LossRet,\,
    \nabla(-\lambda_f \LossFor),\,
    \nabla(-\lambda_m \LossMis)
    \Bigr)$\;
    $\Update(\Delta_P[K_P], g, \eta)$\;
    \If(\tcp*[f]{early stop for projector stage}){$\EvalForget(f_\theta,D_f)$ and $\EvalRetain(f_\theta,D_r)$ satisfy $\Gamma_P$}{
        \textbf{break}\;
    }
}

\BlankLine
\textbf{Continue in Algorithm~\ref{alg:vla_forget_cont}.}\;
\end{algorithm*}

\addtocounter{algocf}{-1}
\begin{algorithm*}[th!]
\caption[]{\textit{VLA-Forget}: Hybrid Unlearning for Vision-Language-Action Models (continued)}
\label{alg:vla_forget_cont}
\DontPrintSemicolon
\SetKwInOut{Input}{Input}
\SetKwInOut{Output}{Output}
\SetKwFunction{InfluenceTriage}{INFLUENCE\_TRIAGE}
\SetKwFunction{RazorSelect}{VISION\_SELECT}
\SetKwFunction{FomulSelect}{LLM\_SELECT}
\SetKwFunction{PCGrad}{PCGRAD}
\SetKwFunction{EvalForget}{EvalForget}
\SetKwFunction{EvalRetain}{EvalRetain}
\SetKwFunction{EvalRobust}{EvalRobust}
\SetKwFunction{TopK}{TopK}
\SetKwFunction{ArgMax}{ArgMax}
\SetKwFunction{Update}{Update}
\SetKwFunction{Enable}{Enable}
\SetKwFunction{Freeze}{Freeze}
\SetKwFunction{Expand}{Expand}

\textbf{Stage 3: Reasoning/action unlearning}\;
$\Freeze(\theta \cup \Delta_V \cup \Delta_P)$; $\Enable(\Delta_L)$ on upper transformer blocks\;
$S_L \leftarrow \FomulSelect(\mathrm{Dec}_{\theta_L}, D_f, D_r, \tau_L)$\;
Optionally add action-token embedding rows / LM-head rows to $S_L$\;
\For{$t = 1$ \KwTo $T_L$}{
    Sample $B_f \sim D_f$, $B_r \sim D_r$, $B_m \sim D_m$\;
    Compute
    \[
    \LossRet(B_r) + \lambda_{\mathrm{feat}}\LossFeat(B_r \cup B_m),\quad
    \LossFor(B_f),\quad
    \LossMis(B_f)
    \]
    $g \leftarrow \PCGrad\Bigl(
    \nabla \LossRet,\,
    \nabla(-\lambda_f \LossFor),\,
    \nabla(-\lambda_m \LossMis)
    \Bigr)$\;
    $\Update(\Delta_L[S_L], g, \eta)$\;
    \If(\tcp*[f]{expand only if forgetting is insufficient}){$\EvalForget(f_\theta,D_f)$ fails $\Gamma_L$}{
        $S_L \leftarrow \Expand(S_L,\ArgMax_{\ell \notin S_L}\mathrm{Sig}(\ell))$\;
    }
    \If(\tcp*[f]{global stopping}){$\EvalForget(f_\theta,D_f)$ and $\EvalRetain(f_\theta,D_r)$ satisfy $\Gamma$}{
        \textbf{break}\;
    }
}

\BlankLine
\textbf{Robustness and deployment audit:}\;
Evaluate closed-loop and offline metrics on retain / forget probes\;
Evaluate quantization robustness under bf16, int8, and int4\;
Evaluate safety metrics (task success, action jerk, gripper toggles, bin saturation)\;
$\mathcal{R} \leftarrow \EvalRobust(f_\theta, D_f, D_r, D_m)$\;

\BlankLine
Merge adapters if desired and return
\[
f_{\theta^\star} \leftarrow f_{\theta_0} \oplus \Delta_V \oplus \Delta_P \oplus \Delta_L
\]
with audit report $\mathcal{R}$\;
\end{algorithm*}


\begin{algorithm*}[th!]
\caption{Helper Procedures for \textit{VLA-Forget}}
\label{alg:vla_forget_helpers}
\DontPrintSemicolon
\SetKwFunction{RazorSelect}{VISION\_SELECT}
\SetKwFunction{FomulSelect}{LLM\_SELECT}
\SetKwFunction{PCGrad}{PCGRAD}
\SetKwProg{Fn}{Function}{}{}

\Fn{\RazorSelect{$M, D_f, D_r, \tau$}}{
    \ForEach{layer or head $l \in M$}{
        $g_l^f \leftarrow \nabla_{\theta_l}\LossFor(D_f)$\;
        $g_l^r \leftarrow \nabla_{\theta_l}\LossRet(D_r)$\;
        $\phi(l) \leftarrow
        \dfrac{\|g_l^f\|_2}{\|\theta_l\|_2+\varepsilon}
        \left(1-\cos(g_l^f,g_l^r)\right)^\alpha$\;
    }
    $K \leftarrow \{l : \phi(l) > \tau\}$\;
    \If{$K = \varnothing$}{
        $K \leftarrow \{\arg\max_l \phi(l)\}$\;
    }
    \Return{$K$}\;
}

\BlankLine
\Fn{\FomulSelect{$B, D_f, D_r, \tau$}}{
    \ForEach{transformer block $l \in B$}{
        $\mathrm{Sig}(l) \leftarrow
        \dfrac{\|\nabla_{\theta_l}\LossFor(D_f)\|_2}
        {\|\nabla_{\theta_l}\LossRet(D_r)\|_2+\varepsilon}$\;
    }
    $S \leftarrow \{l : \mathrm{Sig}(l) > \tau\}$\;
    \If{$S = \varnothing$}{
        $S \leftarrow \{\arg\max_l \mathrm{Sig}(l)\}$\;
    }
    \Return{blocks in $S$ sorted by descending $\mathrm{Sig}(l)$}\;
}

\BlankLine
\Fn{\PCGrad{$\{g_1,\dots,g_n\}$}}{
    Shuffle gradient list\;
    \For{$i=1$ \KwTo $n$}{
        \For{$j=1$ \KwTo $n$}{
            \If{$i \neq j$ \textbf{and} $\langle g_i,g_j\rangle < 0$}{
                $g_i \leftarrow g_i - \dfrac{\langle g_i,g_j\rangle}{\|g_j\|_2^2} g_j$\;
            }
        }
    }
    \Return{$\sum_{i=1}^n g_i$}\;
}
\end{algorithm*}

\paragraph{Algorithm Overview.}
Algorithm~\ref{alg:vla_forget} presents the main \textit{VLA-Forget} unlearning pipeline for Vision-Language-Action models. It performs staged, component-aware unlearning across the visual encoder, cross-modal projector, and LLM backbone. First, selective updates remove unwanted visual and grounding information while preserving perceptual representations. Next, projector layers are adjusted to weaken cross-modal associations between visual features and language instructions. Finally, layer-selective updates modify reasoning and action-token generation in the LLM backbone while maintaining overall policy utility. Algorithm~\ref{alg:vla_forget_helpers} describes the supporting procedures used within this pipeline, including ratio-based module selection, significance-based layer scoring, and PCGrad-based multi-objective optimization to balance forgetting efficacy, perceptual preservation, and reasoning retention.

\subsection{Boundary set construction and mismatch pairing.}
\label{app:mismatch-boundry}
To reduce over-forgetting, we construct the boundary set $D_m$ from near-neighbor samples that are similar to the forget set $D_f$ in scene layout, objects, or instruction wording, but whose behavior should remain unchanged. In practice, for each forget example, we retrieve retain-side samples with high visual or instruction similarity and exclude any instance belonging to the target forget slice, yielding hard non-target examples that lie close to the forget boundary. The mismatch pairing is then formed by aligning each forget sample with its nearest retained counterpart, so that the model is encouraged to diverge on the forgotten mapping while preserving nearby valid behaviors. When an exact same-instruction counterpart is unavailable, we use the closest semantically related retain instruction-scene pair, which provides a stable approximation for enforcing local specificity during unlearning.

\section{Detailed Experimental Setting}
\label{app:sec-exp}

\subsection{Hyper Parameters}
Table~\ref{app-tab:vla_forget_repro} summarizes the key hyper-parameter choices and reproducibility settings used for \textit{VLA-Forget}, including data splits, optimization, unlearning weights, adapter settings, and reporting protocol.

\begin{table*}[t]
\centering
\caption{Hyper-parameter choices and reproducibility details for \textit{VLA-Forget}. Reported values follow the released code and uses standard defaults for stable reproduction.}
\label{app-tab:vla_forget_repro}
\footnotesize
\setlength{\tabcolsep}{4pt}
\begin{tabular}{p{3.2cm}p{3.8cm}p{7.0cm}}
\toprule
\textbf{Category} & \textbf{Setting} & \textbf{Value / Details} \\
\midrule

Model / data 
& Main model 
& \texttt{openvla/openvla-7b}; LoRA-based unlearning in the OpenVLA   \\

&   setups 
& VLA models with vision / projector / reasoning unlearning \\

& Datasets 
& \texttt{OXE}, \texttt{lerobot/pusht\_image} and a   colored-object benchmark \\

& Data scale 
& Up to 512 instances for OpenVLA  ; up to 4000 instances for main  ;   split 5000 / 1200 / 1200 \\

\midrule

Data split 
& Forget fraction 
& 30\% of prompted PushT instances \\

& Train / val / test 
& 70 / 15 / 15 for PushT; separate   train / val / test split \\

& Partition design 
& Forget set $D_f$, retain set $D_r$, and boundary set $D_m$ \\

& Forget target 
& Prompt-based target slice in PushT; \texttt{blue} object target in   benchmark \\

& Evaluation 
& Forget and retain metrics reported on held-out test data \\

\midrule

Optimization 
& Optimizer 
& Adam or AdamW \\

& Base training LR 
& $1\times10^{-3}$ \\

& Unlearning LR 
& $2\times10^{-4}$ \\

& Epochs 
& 6 epochs (main  ), 8 epochs ( ) \\

& Batch size 
& 32 / 64 train-eval for main  ; 64 / 128 for  ; 2 / 2 for OpenVLA   \\

& Update budget 
& 60 steps for OpenVLA  ; steps 60 / 60 / 90 (main) and 100 / 100 / 140 ( ) \\

& Gradient accumulation 
& 8 for OpenVLA   \\

& Gradient clipping 
& Max grad norm $=1.0$ \\

\midrule

Unlearning objective 
& Stage selection 
& Top-$k$ modules: vision $=2$, projector $=2$, reasoning $=3$ or $4$ \\

& Retain weight 
& $\lambda_{\text{retain}}=1.0$ \\

& Forget weight 
& $\lambda_f=0.7$ for OpenVLA  ; $\lambda_f=1.2$ for setting \\

& Mismatch weight 
& $\lambda_m=0.8$ \\

& Feature preservation 
& $\lambda_{\text{feat}}=0.7$ \\

& KL / ratio terms 
& $\beta_{\mathrm{KL}}=0.5$, $\alpha_{\text{ratio}}=1.0$ \\

\midrule

Adapter / precision 
& LoRA setup 
& Rank $r=16$, $\alpha=16$, dropout $=0.05$ \\

& Target modules 
& \texttt{all-linear} in OpenVLA   \\

& Quantization 
& Optional 4-bit quantization; full-precision fallback supported \\

& Compute dtype 
& bf16 when supported, otherwise fp16 \\

& Action setup 
& discrete action heads setups \\

\midrule

Reproducibility 
& Seed 
& 42 for \texttt{random}, \texttt{numpy}, \texttt{torch}, and CUDA \\

& Software 
& PyTorch with \texttt{transformers}, \texttt{datasets}, and \texttt{peft} \\

& Hardware 
& Single modern GPU; CPU fallback only for lightweight tests \\

& Model selection 
& Best checkpoint chosen using validation exact-match accuracy \\

& Reporting 
& Report FC, RC, FAD, RAD, TSR, and SVR; final paper should report mean $\pm$ std over 5 seeds \\

\bottomrule
\end{tabular}
\end{table*}

\subsection{Mathematical Definition of VLA-Forget Metrics}
\label{app:metric}

Let \(D_f=\{(x_i,y_i)\}_{i=1}^{N_f}\) denote the forget set and \(D_r=\{(x_i,y_i)\}_{i=1}^{N_r}\) the retain set, where each input \(x_i=(o_i,s_i)\) contains an observation image \(o_i\) and instruction \(s_i\), and \(y_i=(y_{i,1},\dots,y_{i,T_i})\) is the target action-token sequence produced from the underlying continuous robot action. Since OpenVLA predicts discretized action tokens autoregressively, the natural offline evaluation quantities are token-level and token accuracy on \(D_f\) and \(D_r\). Section~4.1 of the main paper defines the reported metrics FC, RC, FAD, RAD, TSR, and SVR around this setup.

We first define the action-token on any split \(D\in\{D_f,D_r\}\) as
\begin{multline}
\mathrm{CE}_\theta(D)
=
\frac{1}{\sum_{i=1}^{|D|}T_i}
\sum_{i=1}^{|D|}
\sum_{t=1}^{T_i} \\
-\log p_\theta\!\left(y_{i,t}\mid x_i,y_{i,<t}\right).
\end{multline}
Similarly, the token-level action accuracy is

\begin{multline}
\mathrm{Acc}_\theta(D)
=
\frac{1}{\sum_{i=1}^{|D|}T_i}
\sum_{i=1}^{|D|}
\sum_{t=1}^{T_i} \\
\mathbf{1}\!\left[
\arg\max_{v} p_\theta(v\mid x_i,y_{i,<t}) = y_{i,t}
\right].
\end{multline}

In lightweight   evaluations, this accuracy can also be instantiated as exact-match accuracy over the full predicted action token sequence; the same forget/retain deltas are then computed from that accuracy definition.

Using these primitives, the direct offline loss metric is
\[
\mathrm{FC} \;=\; \mathrm{CE}_{\theta_u}(D_f).
\qquad
\]
where \(\theta_u\) denotes the unlearned model. A larger \(\mathrm{FC}\) means the model is less able to reproduce the forgotten action mapping, hence better forgetting.

\noindent \textbf{Retain utility score.}
We define the retain utility score as the negative retain-set cross-entropy,
\[
\mathrm{RC} = -\mathrm{CE}_{\theta_u}(D_r),
\]
so that higher values indicate better retention. Equivalently, one may report the retain improvement relative to the base model,
\[
\mathrm{RC}_{\Delta}
=
\mathrm{CE}_{\theta_0}(D_r)-\mathrm{CE}_{\theta_u}(D_r),
\]
where larger values mean that the unlearned model preserves or improves retain-set behavior relative to the original policy. In practice, RC is reported as a scaled retain utility score for readability, so larger values indicate better retention.

The two accuracy-drop metrics measure change relative to the original pretrained policy \(\theta_0\):
\[
\mathrm{FAD}
=
\mathrm{Acc}_{\theta_0}(D_f) - \mathrm{Acc}_{\theta_u}(D_f),
\]
\[
\mathrm{RAD}
=
\mathrm{Acc}_{\theta_0}(D_r) - \mathrm{Acc}_{\theta_u}(D_r).
\]
Thus, higher \(\mathrm{FAD}\) is better because the forget-set action accuracy should decrease after unlearning, whereas lower \(\mathrm{RAD}\) is better because retained behaviors should change as little as possible. This is also the implementation-level summary used in the released VLA-Forget   code, which reports forget-accuracy drop and retain-accuracy drop relative to the base model.

Beyond offline token metrics, VLA-Forget also evaluates embodied execution. Let \(\mathcal{R}\) be a set of rollout episodes on LIBERO, Open X-Embodiment evaluation slices, or contradiction probes. The task success rate is
\[
\mathrm{TSR}
=
\frac{1}{|\mathcal{R}|}
\sum_{r\in\mathcal{R}}
\mathbf{1}\!\left[\text{task } r \text{ succeeds}\right],
\]
where success is defined by the benchmark-specific completion criterion. Higher \(\mathrm{TSR}\) is better. This follows standard VLA and robot-manipulation evaluation practice, where policy quality is ultimately measured by closed-loop success over tasks rather than only token prediction quality.

Finally, the safety violation rate measures how often the unlearned policy still executes unsafe or disallowed behavior under target prompts, sensitive requests, or contradiction probes:
\begin{multline}
\mathrm{SVR}
=
\frac{1}{|\mathcal{R}_{\mathrm{safe}}|}
\sum_{r\in\mathcal{R}_{\mathrm{safe}}} \\
\mathbf{1}\!\left[\text{episode } r 
\text{ contains a safety violation}\right].
\end{multline}
Typical violations include executing the forbidden action, moving toward the wrong target object, or continuing a visually plausible trajectory under an instruction that should block execution. Hence, lower \(\mathrm{SVR}\) is better. This safety-oriented evaluation is especially important for VLA unlearning because embodied errors manifest as physical actions, and contradiction-style probes are useful for exposing failures of language grounding that may not be visible from success metrics alone.

In summary, \(\mathrm{FC}\) and \(\mathrm{FAD}\) quantify forgetting strength, \(\mathrm{RC}\) and \(\mathrm{RAD}\) quantify retained utility, and \(\mathrm{TSR}\) and \(\mathrm{SVR}\) quantify embodied usefulness and safety. Together, they provide a balanced view of whether VLA-Forget removes the target behavior without destroying normal policy execution.

\section{Additional Results}
\label{app:result}

\subsection{Ablation Study}
\label{app:ablation}

Table~\ref{tab:vla_forget_ablation_appendix_openx} further confirms that the appendix experiments use the \textit{full VLA-Forget configuration}, i.e., joint unlearning over the \textit{vision encoder + projector + language backbone}, together with the staged update schedule. The final configuration is selected with balanced sensitivity settings, using moderate objective weights and a mid-range learning rate (\(\eta=5\times10^{-5}\)), since this setting provides the best overall forget-retain trade-off: it maintains strong forgetting while preserving retain accuracy, reducing action drift, and keeping task success stable, whereas more aggressive settings improve forgetting slightly but noticeably harm retain-side utility.

\begin{table*}[t]
\centering
\caption{Detailed appendix ablation on \textbf{Open X-Embodiment} using \textbf{OpenVLA-7B}. This table extends Table~\ref{tab:vla_forget_ablation} with objective-sensitivity and learning-rate analyses for VLA-Forget. Higher is better for FC, RC, FAD, and TSR; lower is better for RAD and SVR.}
\label{tab:vla_forget_ablation_appendix_openx}
\scriptsize
\setlength{\tabcolsep}{4pt}
\resizebox{\textwidth}{!}{%
\begin{tabular}{llcccccc}
\toprule
\textbf{Group} & \textbf{Setting / Variant} & \textbf{FC $\uparrow$} & \textbf{RC $\uparrow$} & \textbf{FAD $\uparrow$} & \textbf{RAD $\downarrow$} & \textbf{TSR $\uparrow$} & \textbf{SVR $\downarrow$} \\
\midrule

\multicolumn{8}{l}{\textit{A. Component selection (continuation of the main ablation table)}} \\
\midrule
Component & Vision encoder only                           & 85 & 82 & 0.80 & 0.27 & 65 & 12 \\
Component & Projector only                                & 82 & 89 & 0.75 & 0.22 & 75 & 15 \\
Component & Language backbone only                        & 90 & 88 & 0.85 & 0.23 & 74 & 8  \\
Component & Vision + Projector                            & 88 & 86 & 0.84 & 0.24 & 72 & 10 \\
Component & Projector + Language                          & 92 & 90 & 0.86 & 0.22 & 77 & 6  \\
Component & Vision + Language                             & 93 & 87 & 0.89 & 0.25 & 70 & 6  \\
Component & Vision + Projector + Language (full)         & \textbf{93} & \textbf{91} & \textbf{0.88} & \textbf{0.21} & \textbf{78} & \textbf{5} \\
Component & Full, w/o retain-preservation term            & 95 & 80 & 0.93 & 0.30 & 64 & 4 \\
\midrule

\multicolumn{8}{l}{\textit{B. Objective sensitivity around the full VLA-Forget configuration}} \\
\midrule
Objective & Full model, $\lambda_{\text{feat}}=0$ (remove perceptual preservation)      & 94 & 83 & 0.91 & 0.27 & 69 & 5 \\
Objective & Full model, $\lambda_{\text{m}}=0$ (remove mismatch term)                    & 91 & 88 & 0.84 & 0.23 & 75 & 7 \\
Objective & Full model, $\lambda_{\text{f}}=0.5$                                         & 89 & 93 & 0.81 & 0.19 & 79 & 9 \\
Objective & Full model, $\lambda_{\text{f}}=1.0$                                         & 92 & 92 & 0.86 & 0.20 & 78 & 6 \\
Objective & Full model, $\lambda_{\text{f}}=1.5$                                         & 94 & 88 & 0.90 & 0.24 & 74 & 4 \\
Objective & Full model, $\lambda_{\text{m}}=0.5$                                         & 92 & 91 & 0.87 & 0.21 & 78 & 6 \\
Objective & Full model, $\lambda_{\text{m}}=1.0$                                         & \textbf{93} & \textbf{91} & \textbf{0.88} & \textbf{0.21} & \textbf{78} & \textbf{5} \\
Objective & Full model, $\lambda_{\text{m}}=1.5$                                         & 94 & 89 & 0.89 & 0.23 & 76 & 5 \\
Objective & Full model, $\lambda_{\text{feat}}=0.25$                                     & 93 & 89 & 0.89 & 0.23 & 76 & 5 \\
Objective & Full model, $\lambda_{\text{feat}}=0.50$                                     & \textbf{93} & \textbf{91} & 0.88 & \textbf{0.21} & \textbf{78} & \textbf{5} \\
Objective & Full model, $\lambda_{\text{feat}}=1.00$                                     & 91 & 93 & 0.84 & 0.19 & 79 & 7 \\
\midrule

\multicolumn{8}{l}{\textit{C. Learning-rate sensitivity for the full configuration}} \\
\midrule
LR & $\eta = 5\times 10^{-6}$  & 87 & 94 & 0.79 & 0.18 & 80 & 11 \\
LR & $\eta = 1\times 10^{-5}$  & 89 & 93 & 0.82 & 0.19 & 79 & 8  \\
LR & $\eta = 2\times 10^{-5}$  & 91 & 92 & 0.85 & 0.20 & 79 & 7  \\
LR & $\eta = 5\times 10^{-5}$  & \textbf{93} & \textbf{91} & \textbf{0.88} & \textbf{0.21} & \textbf{78} & \textbf{5} \\
LR & $\eta = 1\times 10^{-4}$  & 94 & 88 & 0.90 & 0.24 & 75 & 4  \\
LR & $\eta = 2\times 10^{-4}$  & 95 & 84 & 0.92 & 0.27 & 71 & 4  \\
LR & $\eta = 5\times 10^{-4}$  & 96 & 78 & 0.95 & 0.33 & 62 & 3  \\
\midrule

\multicolumn{8}{l}{\textit{D. Stabilization / schedule controls}} \\
\midrule
Control & Full model, w/o PCGrad                           & 94 & 86 & 0.90 & 0.25 & 73 & 5 \\
Control & Full model, single-stage joint update            & 92 & 87 & 0.86 & 0.24 & 74 & 6 \\
Control & Full model update (ours)                 & \textbf{93} & \textbf{91} & \textbf{0.88} & \textbf{0.21} & \textbf{78} & \textbf{5} \\
Control & Full model, early-stop disabled                  & 95 & 82 & 0.92 & 0.29 & 66 & 4 \\
\bottomrule
\end{tabular}%
}
\end{table*}

\subsection{Safety Case Study}
\label{app:unsafe}

\begin{figure*}[t]
   \includegraphics[width=0.9\linewidth]{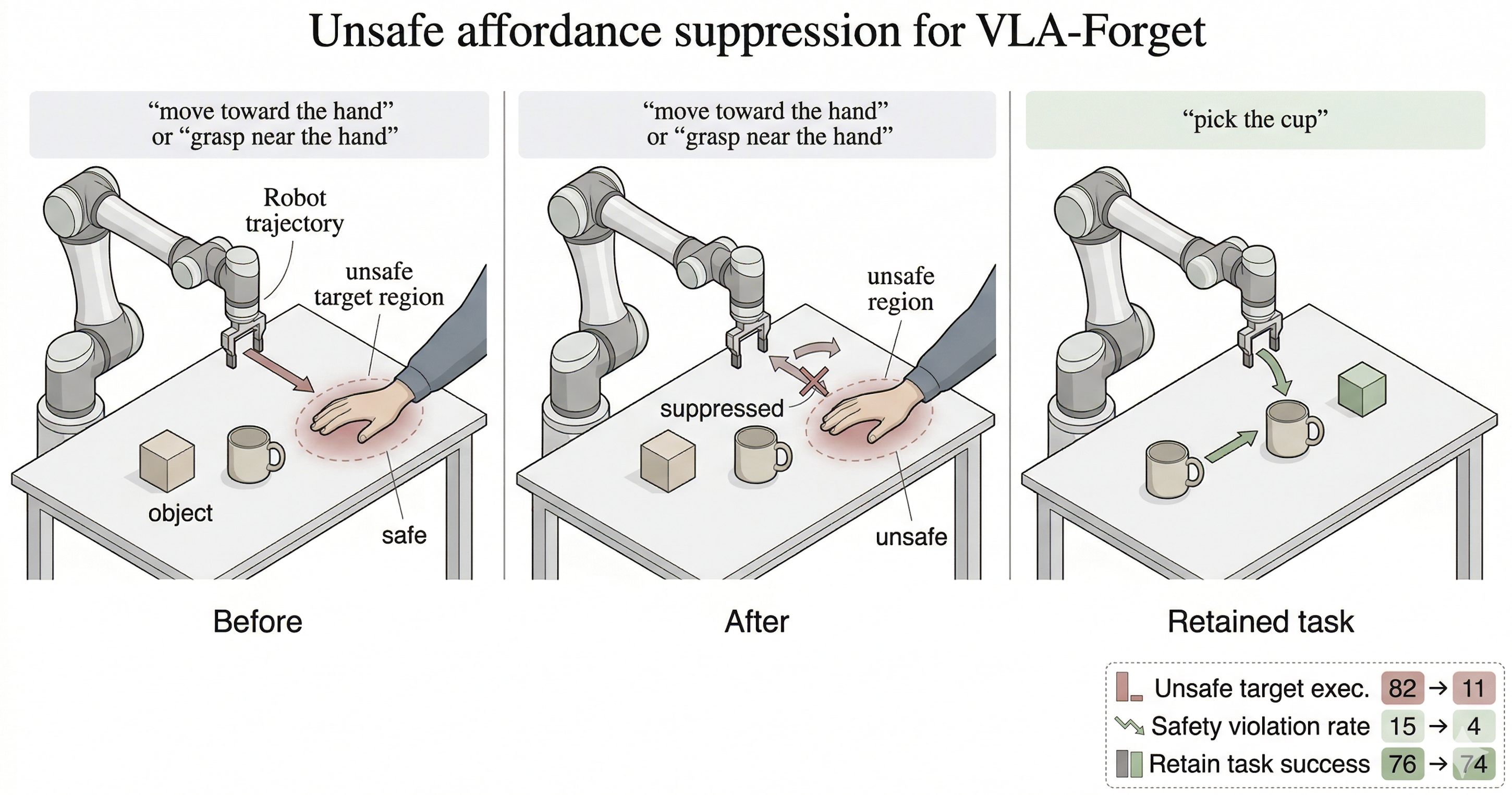}
   \caption{\textbf{Unsafe affordance suppression illustration with retained utility.} Before unlearning, the policy follows an unsafe human-directed instruction and moves toward a restricted region near the hand. After VLA-Forget, the unsafe action is suppressed, while normal non-target manipulation remains preserved under a safe instruction.}
   \label{fig:unsafe_affordance}
\end{figure*}

\noindent
Figure~\ref{fig:unsafe_affordance} provides a qualitative safety case study for VLA-Forget. Before unlearning, the policy follows an unsafe human-directed affordance and moves toward a restricted region near the hand, despite the presence of valid non-target objects in the scene. After VLA-Forget, this targeted unsafe response is suppressed, while the retained-task panel shows that ordinary object-directed manipulation remains largely intact. This behavior is consistent with the paper’s central claim that VLA unlearning should remove a specific unsafe instruction-to-action association, rather than broadly degrade perception or manipulation ability, thereby yielding a more useful and deployment-relevant safety correction. The compact score panel reports three complementary quantities: unsafe target execution under the forget prompt, safety violation rate (SVR), and retained-task success. Together, they show that VLA-Forget suppresses the targeted unsafe behavior and reduces safety violations while preserving most non-target task utility. 
The iterative unlearning procedure in \textsc{VLA-Forget} enables more precise and controlled removal of targeted behavior than a single-step gradient nudge such as G-Drift~\cite{ranjan2026g}. VLA-Forget could potentially adopt the reasoning framework mention in KORAL~\cite{akewar2026koral}. It also operates directly on the deployed VLA policy, without requiring model restructuring or external retrieval documents as in CatRAG~\cite{ranjan2026catrag,ranjan2026position}.

\end{document}